\documentclass[letterpaper, 10 pt, conference]{ieeeconf}  

\IEEEoverridecommandlockouts                              

\usepackage{graphics} 
\usepackage{epsfig} 
\usepackage{subfig}
\usepackage{amsmath} 
\usepackage{amssymb}  
\usepackage{booktabs} 
\usepackage{cite}

\usepackage{algorithm}  
\usepackage{algorithmic} 

\title{\LARGE \bf
Regeneration and Joining of the Learned Motion Primitives for Automated Vehicle Motion Planning Applications
}

\author{Boyang Wang$^{1,2}$,~\IEEEmembership{Student Member,~IEEE}, Jianwei Gong$^{1}$,~\IEEEmembership{Member,~IEEE}, Wenli Liang$^{1}$, Huiyan Chen$^{1}$
\thanks{*This work was supported by the National Natural Science Foundation of China (No.91420203 and No.61703041), and the paper is also funded by International Graduate Exchange Program of Beijing Institute of Technology}
\thanks{$^{1}$ All the authors are with the School of Mechanical Engineering, Beijing Institute of Technology, Beijing, China, 100081
        {\tt\small gongjianwei@bit.edu.cn}}%
\thanks{$^{2}$ Boyang Wang is also with the Interactive Digital Human group of CNRS-UM LIRMM, UMR5506, Montpellier, France, 34095.
        {\tt\small wbythink@hotmail.com}}%
}

\begin{document}

\maketitle
\thispagestyle{empty}
\pagestyle{empty}

\begin{abstract}

How to integrate human factors into the motion planning system is of great significance for improving the acceptance of intelligent vehicles. Decomposing motion into primitives and then accurately and smoothly joining the motion primitives (MPs) is an essential issue in the motion planning system. Therefore, the purpose of this paper is to regenerate and join the learned MPs in the library. By applying a representation algorithm based on the modified dynamic movement primitives (DMPs) and singular value decomposition (SVD), our method separates the basic shape parameters and fine-tuning shape parameters from the same type of demonstration trajectories in the MP library. Moreover, we convert the MP joining problem into a re-representation problem and use the characteristics of the proposed representation algorithm to achieve an accurate and smooth transition. This paper demonstrates that the proposed method can effectively reduce the number of shape adjustment parameters when the MPs are regenerated without affecting the accuracy of the representation. Besides, we also present the ability of the proposed method to smooth the velocity jump when the MPs are connected and evaluate its effect on the accuracy of tracking the set target points. The results show that the proposed method can not only improve the adjustment ability of a single MP in response to different motion planning requirements but also meet the basic requirements of MP joining in the generation of MP sequences.

\end{abstract}

\section{Introduction}

The integration of the human factors extracted from the collected driving data into the existing automated driving system has a significant contribution to improving the performance of the automated vehicles\cite{martinez2018driving,olaverri2016human,duan2017driver,li2017estimation,guo2018humanlike}. The motion planning system can generate appropriate trajectories according to the driving situations and is an important part of the automated driving system\cite{gonzalez2016review,paden2016survey}. One of the effective methods is to decompose the motion into primitives, and transform the motion planning problem into MP generation, selection and joining problems\cite{werling2010optimal,chu2012local,hu2018dynamic}. In order to further improve the practicability of the learned MP library for motion planning tasks, this paper mainly solves the following two fundamental issues: one is how to expand the adjustment ability further when the single MP is re-generated, and the other one is how to solve the smooth transition problem when MPs are connected.

\begin{figure*}[t]
\centering
\subfloat[The structure of the learned MP library.]{
\begin{minipage}[t]{0.45\textwidth}
\centering
\includegraphics[scale=0.25]{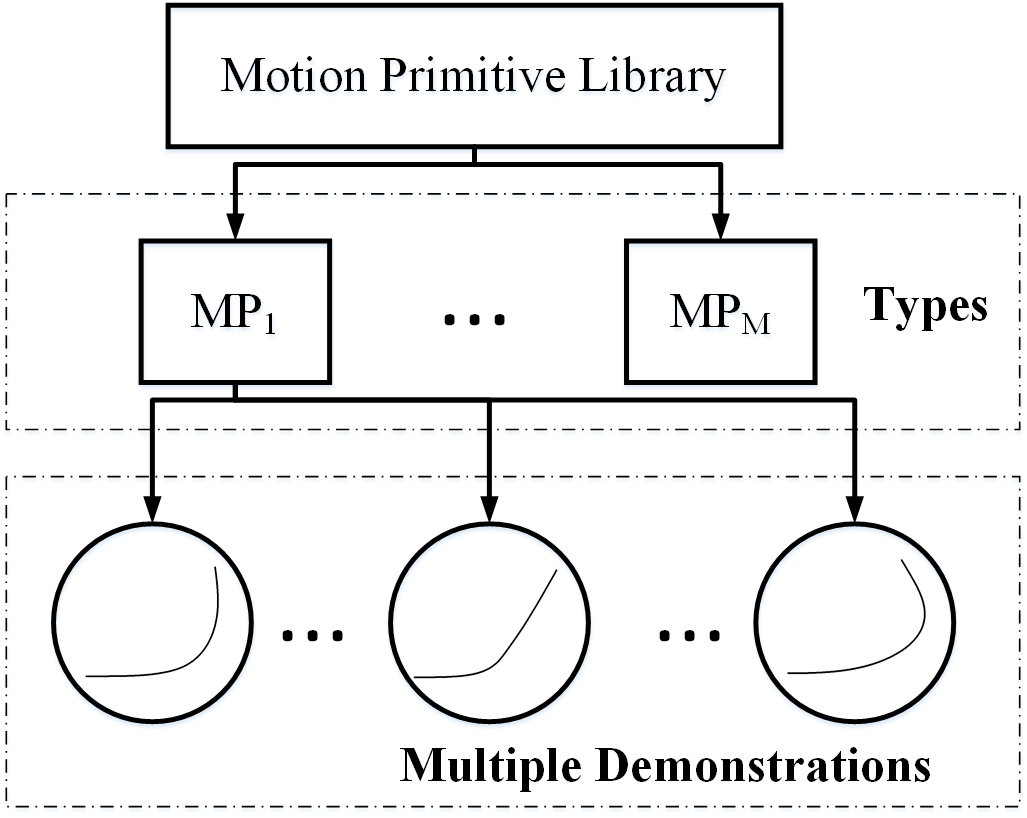}
\end{minipage}
\label{fig１(a)}
}
\hspace{20pt}
\subfloat[MP regeneration and joining process.]{
\begin{minipage}[t]{0.45\textwidth}
\centering
\includegraphics[scale=0.25]{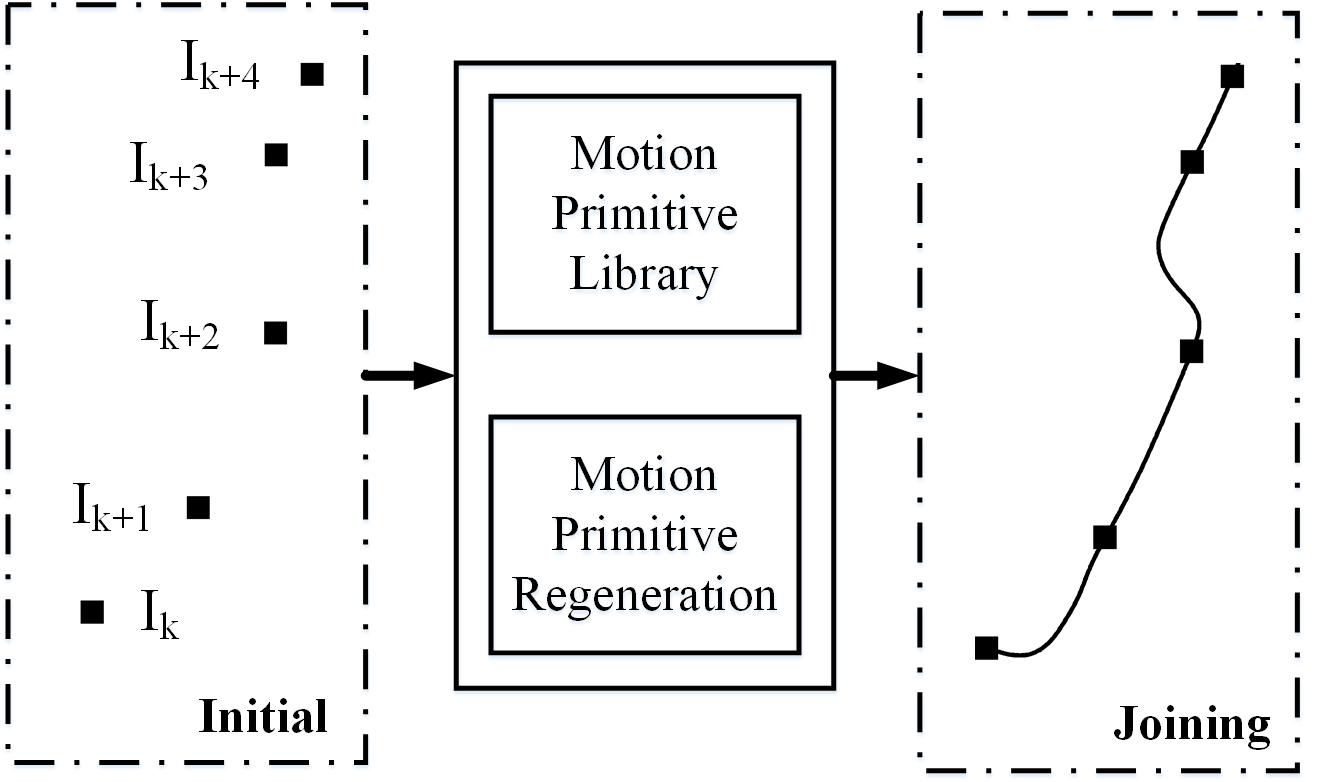}
\end{minipage}
\label{fig１(b)}
}

\caption{The overall flow of our framework: First, we proposed a modified MP representation method to represent the multiple demonstrated trajectories under the same type in the MP library. Second, we completed the regeneration of a single MP and the joining of multiple MPs according to the given initial conditions.}
\label{fig１}
\end{figure*}

\par In recent years, many methods have been applied to solve the problem of motion planning for automated vehicle in various driving situations. These methods mainly include the graph search-based planners, sampling-based planners, interpolating curve planners, and discrete optimization approaches\cite{gonzalez2016review}. Among these methods, the discrete optimization method uses a limited trajectory set to adapt to the changing driving environment and improves the efficiency of the algorithm by reducing the solution space. Therefore, it has been widely used in automated driving systems in recent years. Werling et al.\cite{werling2010optimal} generated a trajectory set with several lateral offsets from a fixed-baseline based on quintic polynomials and then ensures a smooth transition between the trajectory segments through cost function. The proposed trajectory generation method was applied to achieve static obstacles avoidance in\cite{chu2012local} and moving obstacles avoidance in\cite{hu2018dynamic}. However, the candidate trajectory set generated by the algorithm mentioned above is not based on the real driving data and does not consider the human factors in the motion planning system.

\par There are many ways to represent the human manipulation skills in the field of robotics, including spline-based via-point models, Hidden Markov Models (HMMs), Gaussian Mixture Models (GMMs) and DMPs\cite{mulling2013learning}. Among these methods, DMP has attracted many researchers with its strong robustness and adaptability\cite{ijspeert2013dynamical}. The application of DMP for human behavior representation includes biped walking\cite{huang2019learning}, ball hitting\cite{mulling2013learning}, reaching a position\cite{gavspar2018skill}, bimanual manipulation of clothes\cite{colome2018dimensionality}, etc. The collection of human demonstration data is often done by a motion capture device or by recording the joint angle feedback values of the guided robot arm. After learning one or several task-related MPs through the demonstration data, the motion planning system selects and regenerates the MPs according to the requirements. The above human skills collection experiments are often limited to specific repetitive scenes, but the driving data collection experiments are often carried out in a more general traffic environment. How to deal with the relative position data obtained from the general driving experiment, rather than the absolute joint angle data obtained from a specific scene, is a major challenge in this paper. 

\par In the analysis and utilization of the collected vehicle trajectory, Zhao et al.\cite{zhao2017road} developed an on-road vehicle trajectory collection system to collect the driving data. A trajectory library for representing the driver's lane change behavior was established by analyzing the collected driving data\cite{yao2017road,xu2018naturalistic}. The library was then used to match the set of candidate trajectories generated by the motion planning system to achieve generation of human-like lane-changing trajectories\cite{he2018human}. Schnelle et al.\cite{schnelle2017driver} proposed a trajectory generation method which can fit the road geometry and achieve the personalized driving, and validated the algorithm through lane change and double lane change scenarios. The authors further extended the method to the field of lane exchanging, avoiding collisions and achieving the integration of driving habits\cite{wang2017framework}. Although the methods mentioned above can implement human-like trajectory planning, they are specific to certain scenarios and therefore lack generality.

\par In this paper, we propose a modified trajectory representation method which can enhance the adjustable ability of a single MP and achieve the joining between multiple MPs in the library. In the authors' early work\cite{wang2018learning1,wang2018learning}, the driving database under general traffic scene has been established, and different types of MPs have been extracted and identified to form the MP library (Fig.\ref{fig１(a)}). Based on the previous work, this paper uses SVD to realize the further processing of multiple demonstrated trajectories in the same type, which not only extracts the main shape components but also reduces the dimension of the shape adjustment parameter set during regeneration. In addition, we implement the nonlinear function, goal function and canonical system association of each independent MP in the MP sequence through the modified representation method. Through the above association, we have completed a smooth connection between the MPs in the sequence (Fig.\ref{fig１(b)}). 

\par The main contributions of this paper are shown as follows:

\begin{itemize}		

\item Providing a decomposition method for the shape representation parameters of the same type of MPs, which simplifies and constrains the shape adjustment during regeneration.

\item Solving the problem of MPs joining by utilizing the modified representation method to achieve the association between independent MPs in the sequence.

\end{itemize}	

\par The remains of the paper are organized as follows. Section \uppercase\expandafter{\romannumeral2} details the issues to be solved in this article. Section \uppercase\expandafter{\romannumeral3} presents the modified representation method and the joining method which are used to solve the problem. Section \uppercase\expandafter{\romannumeral4} shows the results of MP regeneration and joining. Finally, the conclusions are given in Section \uppercase\expandafter{\romannumeral5}.

\section{Problem Statement}

According to a learned MP library \begin{math}  \boldsymbol{M} = \{ \boldsymbol{m}_1, \ldots ,\boldsymbol{m}_M\}  \end{math}, the goal of this paper is to further separate the shape representation parameters of the same type of MPs, and generate the MP sequence \begin{math}  \boldsymbol{R} = \{ \boldsymbol{r}_1, \ldots ,\boldsymbol{r}_I\}  \end{math} based on the initial condition \begin{math}  \boldsymbol{I} = \{ \boldsymbol{I}_1, \ldots ,\boldsymbol{I}_K\}  \end{math}.

\par The main parameters used in this paper are defined as follows:

\begin{itemize}		

\item \begin{math}  \boldsymbol{m}_m(t) = {[{x_m}(t),{y_m}(t)]^{\rm{T}}} \in {\mathbb{R}^{2 \times 1}}  \end{math} is the definition of one MP point at time $t$, where $x_m$ and $y_m$ are the longitudinal and lateral position value in the $x_moy_m$ coordinate. The initial position is the coordinate origin, and the initial heading direction is consistent with the positive direction of the $x_m$-axis.

\item \begin{math}  \boldsymbol{\omega} _m = {[\boldsymbol{\omega} _x,\boldsymbol{\omega} _y]^{\rm{T}}}  \end{math} is the inherent attribute set of the selected MP which remains constant during the adjustment of the MP sequence generation, where $\boldsymbol{\omega} _x$ and $\boldsymbol{\omega} _y$ are the basic shape characterization parameters of the longitudinal and lateral trajectory shapes of the chosen MP type.

\item \begin{math} \boldsymbol{\gamma} _m = {[\boldsymbol{b},\boldsymbol{g},T,\boldsymbol{s}]^{\rm{T}}} \end{math} is the adjustment parameter set of the selected MP which can be changed during the generation process to satisfy different motion planning task requirements, where $\boldsymbol{b}$ is the initial position, $\boldsymbol{g}$ is the goal position $T$ is the time duration and $\boldsymbol{s}$ is the fine tuning parameter set for the trajectory shape which allows us to make further adjustments based on the basic shape of the selected type.

\item \begin{math} \boldsymbol{I}_k = {[id,T,{x_{init}},{y_{init}},{x_{g}},{y_{g}}]^{\rm{T}}} \in {\mathbb{R}^{6 \times 1}}  \end{math} defines the initial condition for the generation of MP sequence, where $id$ is the identification information of the ${k^{{\rm{th}}}}$ MP in the sequence, $T$ is the corresponding time duration of the selected MP, $x_{init}$ and $y_{init}$ are the initial position, $x_{g}$ and $y_{g}$ are the goal position of the ${k^{{\rm{th}}}}$ MP.

\item \begin{math} \boldsymbol{r}_i(t) = {[x(t),y(t)]^{\rm{T}}} \in {\mathbb{R}^{2 \times 1}}  \end{math} is the definition of the generated MP sequence point at time $t$, $x(t)$ and $y(t)$ are the generated position value in the $xoy$ coordinate. The $xoy$ coordinate system is consistent with the $x_moy_m$ coordinate system of the first MP in the sequence.

\end{itemize}	

\par The proposed method in this paper is trying to solve the following two challenges: First, how to extract the basic shape parameter set $\boldsymbol{\omega} _m$ and the fine-tuning shape parameter set $\boldsymbol{s}$ from multiple demonstration trajectories which are interrelated but not identical in each type of MP in the library. Second, how to ensure a smooth transition of the velocity at the switching points in the MP sequence given by the initial condition $\boldsymbol{I}$.

\section{Methodology}

\subsection{Modified Representation Method of MPs} 

The original DMP algorithm only has the ability to represent a single demonstration trajectory\cite{wang2018learning}, which is defined as

\begin{gather}
{\dot z_m} =  - \tau {\alpha _z}{z_m}     \label{equ1} \\
{\boldsymbol{{\dot v}}_m} = \tau {\alpha_m}({\beta_m}(\boldsymbol{g} - \boldsymbol{d}_m) - \boldsymbol{v}_m) + \tau (\boldsymbol{g} - \boldsymbol{b})f({z_m})            \label{equ2} \\
\boldsymbol{{\dot d}}_m = \tau \boldsymbol{v}_m   \label{equ3} \\
f({z_m}) = \frac{{\sum\nolimits_{n = 1}^N {{\omega _n}{\psi _n}({z_m})} {z_m}}}{{\sum\nolimits_{n = 1}^N {{\psi _n}({z_m})} }} \label{equ4}
\end{gather}

\noindent where $\boldsymbol{d}_m=[{x_m},{y_m}]$. $\boldsymbol{v}_m=[{\dot x_m},{\dot y_m}]$, $\boldsymbol{g}=[{x_m}(T),{y_m}(T)]$ is the goal position of the selected MP and $\boldsymbol{b} = [{x_m}(0),{y_m}(0)]$ is the initial position. ${\psi _n}({z_m}) = exp( - {p_n}{({z_m} - {\mu _n})^2})$ is the Gaussian basis function, $\omega _n$ is an element in the shape representation parameter set, and $N$ is the total number of the elements in the parameter set. The constants $\alpha_m$ and $\beta_m$ are pre-defined to ensure the spring-damper system is critically damped. and $\tau$ is the time scaling parameter. 

\par In order to enable the proposed method to represent multiple trajectories of the same type in the MP library and provide an interface for subsequent MPs joining, we propose a modified DMP method. The modified canonical system is defined as

\begin{gather}
{{\dot z}_m} =  - \frac{{{\alpha _z}{e^{{\alpha _z} \cdot (\tau T - t)/\Delta t}}}}{{{{(1 + {e^{{\alpha _z} \cdot (\tau T - t)/\Delta t}})}^2}}} \label{equ5}
\end{gather}

\noindent where $T$ is the duration of the MP, ${\Delta t}$ is the sampling interval. Because multiple demonstration trajectories in the same type of MP do not necessarily have the same duration, in order to ensure that the vectors after resampling have the same size, the sampling interval $\Delta t = T/100$. 

\par We introduce the goal function to replace the goal position value to avoid the acceleration jump caused by the sudden change of the goal position at the switching point. The modified form is proposed as follows:

\begin{gather}
\boldsymbol{\dot v}_m = \tau {\alpha _m}({\beta _m}(\boldsymbol{r}_m - \boldsymbol{d}_m) - \boldsymbol{v}_m) + f(t,{z_m})  \label{equ6} \\
\boldsymbol{\dot d}_m = \tau \boldsymbol{v}_m  \label{equ7} \\
\tau \mathop {\boldsymbol{r}_m}\limits^ \cdot   = \left\{ \begin{array}{l}
\frac{{\Delta t}}{T}(\boldsymbol{g}_m - \boldsymbol{b}_m),{\rm{if}} {\ }t \le T\\
0, {\ }{\rm{else}}
\end{array} \right.   \label{equ8}
\end{gather}

\par The nonlinear function $f(t,{z_m})$ is the core factor in representing the shape transformation of trajectories. In our modified representation algorithm, the definition of the nonlinear transformation function for one-DOF is as follows:

\begin{gather}
f(t,{z_m}) = {\alpha _w}\left( {\sum\limits_{j = 1}^J {\frac{{\sum\nolimits_{i = 1}^N {{\omega _{ji}}{\psi _{ji}}(t){s_j}} }}{{\sum\nolimits_{i = 1}^N {{\psi _{ji}}(t)} }}} } \right){z_m}   \label{equ9}
\end{gather}

\noindent where ${\psi _{ji}}(t) = exp( - {p_{ji}}{(\frac{t}{{\tau T}} - {\mu _{ji}})^2})$ is the Gaussian basis kernel of the modified DMP with $\mu_{ji}$ represents the center and $p_{ji}$ represents the bandwidth. $\boldsymbol{\omega}  = [\boldsymbol{\omega} _1, \cdots ,\boldsymbol{\omega} _J]$ and $\boldsymbol{s} = [\boldsymbol{s}_1, \cdots ,\boldsymbol{s}_J]$ are the parameter matrices of one-DOF data in the inherent attribute set $\boldsymbol{\omega} _m$ and fine tuning parameter set $\boldsymbol{s}$ respectively. These two parameter sets $\boldsymbol{\omega}$ and $\boldsymbol{s}$ are combined to be used as the representation parameters of the trajectory shape.

\subsection{Learning Process of MP Parameter Set} 

The training data is a set of multiple demonstration trajectories with a similar shape obtained by the probabilistic segmentation algorithm. The learning process consists of two main processes: The first is to extract the basic shape representation data $\boldsymbol{D}_{baiss}$ from multiple demonstrations by using the SVD algorithm. The second is to obtain the inherent attribute set $\boldsymbol{\omega} _m$ by minimizing the errors between representation and demonstration. Since the training process of the longitudinal and lateral parameters is the same, we will only introduce the training process of the longitudinal part.

\par Assuming that there are $Q$ demonstration trajectories in the selected MP type, data resampling ensures that each segmented trajectory has $C$ sampling points, so all longitudinal trajectory data in this type can be represented as the same size of the vector $\boldsymbol{x}_q^{dem} \in {\mathbb{R}^{C \times 1}}$.

\par We can calculate the specific value to be represented by the nonlinear function in each demonstration trajectory by applying the following equation:

\begin{gather}
f_{dem}^m(t,z_m) = {\ddot x_q}(t)/{\tau ^2} - {\alpha _y}({\beta _y}(r - {x_q}(t)) - {\dot x_q}(t)) \label{equ10}
\end{gather}

\par The data set $\boldsymbol{F}_{dem} = [\boldsymbol{f}_{dem}^1, \cdots ,\boldsymbol{f}_{dem}^Q]^{\rm{T}} \in {\mathbb{R}^{Q \times C}}$ will be obtained after calculating all the demonstration trajectories of the selected type. Then we apply the SVD to decompose the $Q \times C$ matrix $\boldsymbol{F}_{dem}$, and the decomposed matrix is represented as follows:

\begin{gather}
\boldsymbol{{F}}_{dem} = {\boldsymbol{U\Sigma }}{{\boldsymbol{V}}^{\rm{T}}} \approx {\boldsymbol{s}}{{\boldsymbol{D}}_{basis}}  \label{equ11}
\end{gather}

\noindent where the basic shape representation data ${\boldsymbol{D}_{basis}} = {[\boldsymbol{D}_{basis}^1, \cdots ,\boldsymbol{D}_{basis}^J]^{\rm{T}}} \in {\mathbb{R}^{J \times C}}$ is the first $J$ columns of ${\boldsymbol{\Sigma }}{{\boldsymbol{V}}^{\rm{T}}}$. The fine tuning parameter set $\boldsymbol{s} = [{s^1}, \cdots .{s^Q}]^{\rm{T}} \in {\mathbb{R}^{Q \times J}}$ is the first $J$ columns of $\boldsymbol{U}$. The dimension $J$ is determined by the singular value spectrum. In general, the value of $J$ is much smaller than $Q$ ($J \ll Q$).

\par The learning of inherent attribute set ${\boldsymbol{\omega} _x} = [{\boldsymbol{\omega} _1}, \cdots ,{\boldsymbol{\omega} _J}]^{\rm{T}} \in {\mathbb{R}^{J \times N}}$ is to solve the optimization problem as shown in Eq.\ref{equ12}, and the problem can be solved by a regression algorithm mentioned in the previous study\cite{wang2018learning}.

\begin{gather}
\omega _j^ *  \leftarrow \arg \mathop {\min }\limits_{{\omega _j}} \sum\limits_{c = 1}^C {\left( {D_{basis}^j({t_c}) - \frac{{\sum\nolimits_{i = 1}^N {{\omega _{ji}}{\psi _{ji}}({t_c})} }}{{\sum\nolimits_{i = 1}^N {{\psi _{ji}}({t_c})} }}{z_m}} \right)}  \label{equ12}
\end{gather}

\par Through the above process, the ${\boldsymbol{\omega} _m} = {[{\omega _x},{\omega _y}]^{\rm{T}}}$ is learned to represent the separated basic shape data.

\subsection{Generating the MP Sequence} 

\par The simple joining method is to execute the next MP at the end of the previous MP, but this simple method only has a simple position correlation capability, i.e., the termination position of the previous MP is used as the starting position of the next MP, but the method does not have the velocity correlation ability in the process of generating MP sequences.

\par Therefore, we propose a novel joining method based on the modified MP representation algorithm to solve the problem of smooth transition. In the proposed joining method, we treat the MPs in the generated sequence as a whole rather than separate individual. A single set of overlapping kernels is established to construct the connections between different kinds of MPs in the sequence.

\par The centers $\boldsymbol{\mu} '$ and width $\boldsymbol{p}'$ of the Gaussian kernels are regenerated as follows:

\begin{gather}
\mu _k^i = \left\{ \begin{array}{l}
\frac{{{T_1} \cdot (i - 1)}}{{T' \cdot (N - 1)}},{\rm{  if {\ } k = 1}}\\
\frac{{{T_k} \cdot (i - 1)}}{{T' \cdot (N - 1)}} + \frac{1}{{T'}}\sum\limits_{l = 1}^{k - 1} {{T_l},{\rm{ {\ } else}}}
\end{array} \right.  \label{equ13}  \\
\boldsymbol{\mu} ' = \mu _1^1,...,\mu _1^N;...;\mu _K^1,...,\mu _K^N  \label{equ14}  \\
\bar p_k^i = \frac{{p_k^i \cdot T'}}{{{T_k}}}  \label{equ15}  \\
\boldsymbol{p}' = \bar p_1^1,...,\bar p_1^N;...;\bar p_K^1,...,\bar p_K^N  \label{equ16}
\end{gather}

\noindent where $K$ is the total number of MPs in the generated sequence, and $T_k$ is the time duration of ${k^{{\rm{th}}}}$ MP in the sequence which is given by the initial condition $\boldsymbol{I}_k$. $T' = \sum\limits_{k = 1}^K {{T_k}}$ is the total time duration of the regenerated MP sequence. The inherent attribute set $\omega '$ is a simple combination of each parameter set learned from the demonstration. 

\begin{gather}
\boldsymbol{\omega} ' = \omega _1^1,...,\omega _1^N;...;\omega _K^1,...,\omega _K^N  \label{equ17}
\end{gather}

\par According to the given initial condition $\boldsymbol{g_k} = [{x_g^k},{y_g^k}]$ and $\boldsymbol{b_k} = [{x_{init}^k},{y_{init}^k}]$, the goal function of the MP sequence is defined as:

\begin{gather}
\tau \dot r' = \left\{ \begin{array}{l}
\frac{{\Delta {t_k}}}{{{T_k}}}({g_k} - {b_k}),{\rm{  if {\ }}}\sum\limits_{l = 1}^{k - 1} {{T_l} \le t \le \sum\limits_{l = 1}^k {{T_l}} } \\
0,{\rm{  {\ } otherwise}}
\end{array} \right.   \label{equ18}
\end{gather}

\par The canonical system $z'$ is defined as 

\begin{gather}
\dot z' =  - \frac{{{\alpha _z}{e^{{\alpha _z} \cdot (\tau T' - t)/\Delta {t_k}}}}}{{{{(1 + {e^{{\alpha _z} \cdot (\tau T' - t)/\Delta {t_k}}})}^2}}}     \label{equ19}
\end{gather}

\par Besides, since the coordinate system of a single MP is $x_moy_m$, and the coordinate system of the MP sequence system is $xoy$, the coordinate transformation is required in the process of generating the MP sequence.

\begin{gather}
{{\dot v'}_x} = \tau {\alpha _m}({\beta _m}({{r'}_x} - {{d'}_x}) - {{v'}_x}) + {F_{tran\_x}}  \label{equ20}   \\
{F_{tran\_x}} = \cos {\delta _j} \cdot {f_x}(t,z) - sin{\delta _j}{f_y}(t,z)  \label{equ21} \\
\dot x = \tau {v_x}  \label{equ22}   \\
{{\dot v}_y} = \tau {\alpha _m}({\beta _m}({r_y} - {d_y}) - {v_y}) + {F_{tran\_y}} \label{equ23}   \\
{F_{tran\_y}} = \sin {\delta _j} \cdot {f_x}(t,z) + \cos {\delta _j}{f_y}(t,z) \label{equ24}   \\
\dot y = \tau {v_y}    \label{equ25} 
\end{gather}

\noindent where ${\delta _j}$ is the course angle at the end of the previous MP which is defined in the coordinate system $xoy$. $d_x$, $d_y$ are the position values and $v_x$, $v_y$ are the velocity values in the coordinate system $xoy$ of the generated MP sequence. ${f_x}(t,z)$ and ${f_y}(t,z)$ are the nonlinear functions which are based on the redefined overlapping Gaussian kernels. 

\par Based on the given initial condition $\boldsymbol{I}$ and Eq.\ref{equ13} to Eq.\ref{equ25}, we can achieve a smooth connection between multiple MPs in the generated MP sequence.

\section{Experiments and Results}

\subsection{Data Collection and Preprocessing} 

The data set used in this paper is the same as the data set used in the previous study of MP representation and segmentation\cite{wang2018learning}. The established MP library which contains the segmented and classified observed trajectory data of a single human driver is selected as the training dataset.

\par The position values in the representation dataset $\boldsymbol{m}_m$ are calculated by integrating the vehicle speed and course angle acquired at equal time intervals. The vehicle speed is collected by the CAN bus, the course angle is obtained from the integrated navigation unit, and the time interval of the data collection system is 100ms. In order to maintain the consistency of the data dimension, multiple observed trajectories included in each type of MP need to have the same number of sampling points. Therefore, before the MP training process, the trajectory data need to be resampled so that each trajectory has 100 sampling points.

\subsection{Evaluation of Typical MP Representation and Regeneration} 

\begin{figure}
\centering
\subfloat[The type of sharp turn MP]{
\begin{minipage}[t]{0.45\textwidth}
\centering
\includegraphics[scale=0.2]{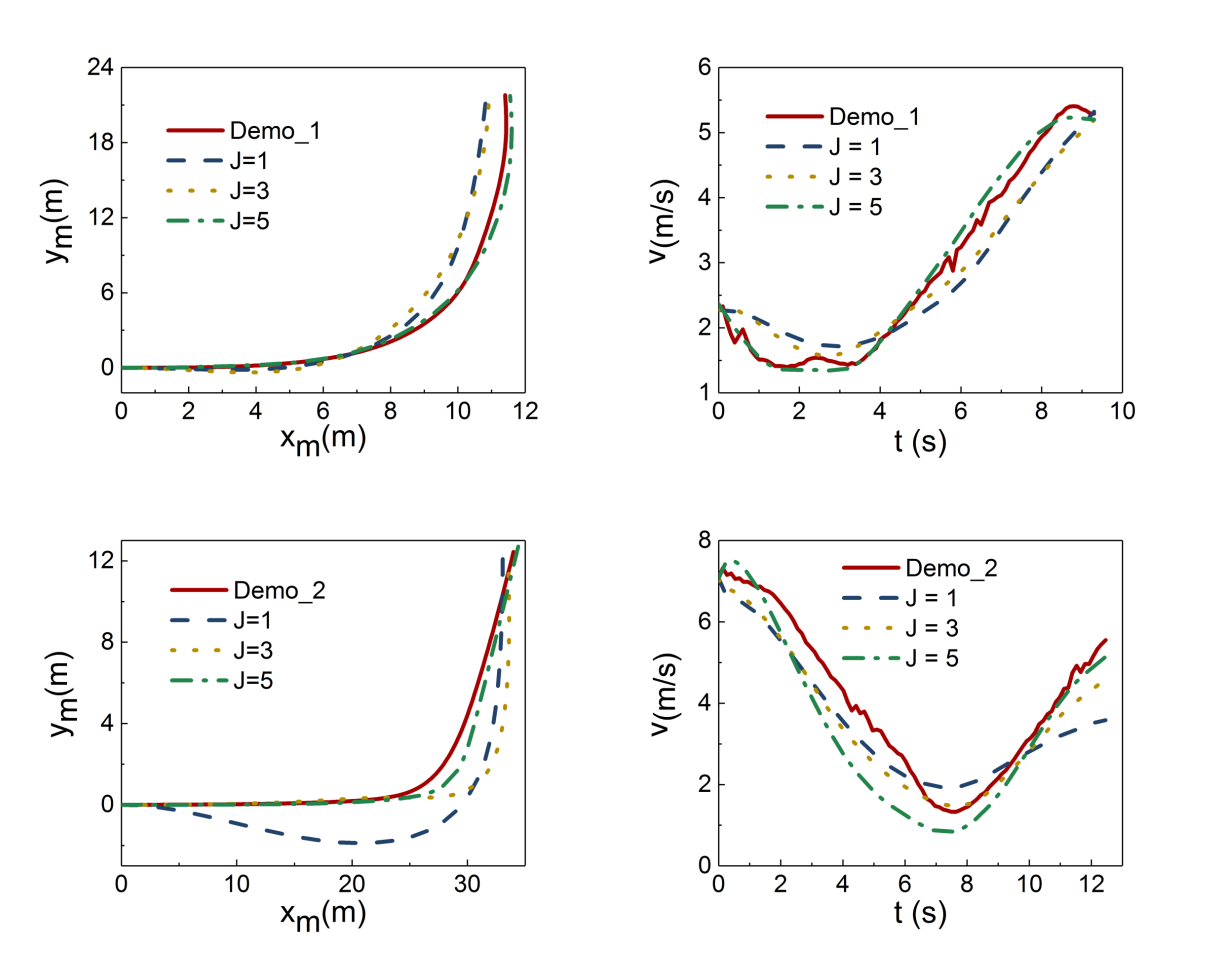} 
\end{minipage}
\label{fig2(a)}
}

\subfloat[The type of lane changing MP]{
\begin{minipage}[t]{0.45\textwidth}
\centering
\includegraphics[scale=0.2]{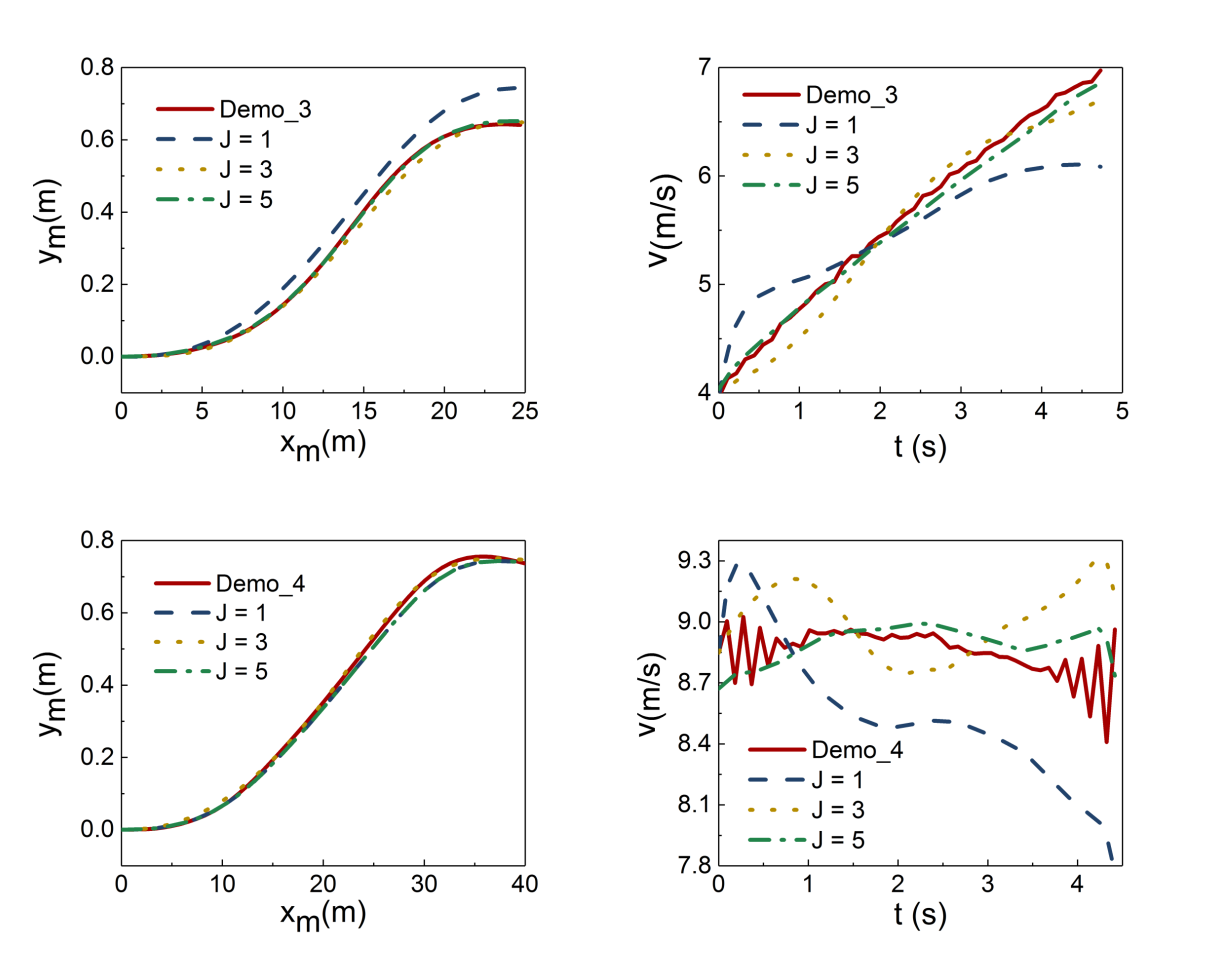} 
\end{minipage}
\label{fig2(b)}
}
\caption{This figure illustrates the representation accuracy of the proposed representation method and the influence of the number of fine-tuning parameters $\boldsymbol{s}$. In each selected MP type, we selected two different demonstration trajectories of the same type to analyze the effect.}
\label{fig2}
\end{figure}

The purpose of the MP training is to represent the driver's various typical driving behaviors, including lane keeping, lane changing and other driving trajectory segments with different steering correction levels at different speeds. Considering that the MPs demonstrated here should be the most common driving behavior and contain the typical longitudinal-lateral cooperation relationship, the sharp turn MP at low speed and lane changing MP at medium speed were selected from the established MPs library.

\begin{table}
 \centering
 \caption{\label{tab1}This table lists the average deviations in position and velocity when the number of fine-tuning parameters $\boldsymbol{s}$ changed. Type 1 refers to the sharp turn, and type 2 refers to the lane changing. }
 \begin{tabular}{c c c c}
  \toprule
  Type & $J$ & ${\Delta \bar d}$(m) & ${\Delta \bar v}$(m/s) \\ 
  \midrule
  1 & 1 & 2.12 & 0.43  \\
  1 & 3 & 2.09 & 0.41 \\
  1 & 5 & 1.49 & 0.29 \\
  2 & 1 & 1.78 & 0.41  \\
  2 & 3 & 1.72 & 0.32 \\
  2 & 5 & 1.46 & 0.25 \\
  \bottomrule
 \end{tabular}
\end{table}

\begin{figure*}
\centering
\subfloat[The type of sharp turn MP]{
\begin{minipage}[t]{0.9\textwidth}
\centering
\includegraphics[scale=0.1865]{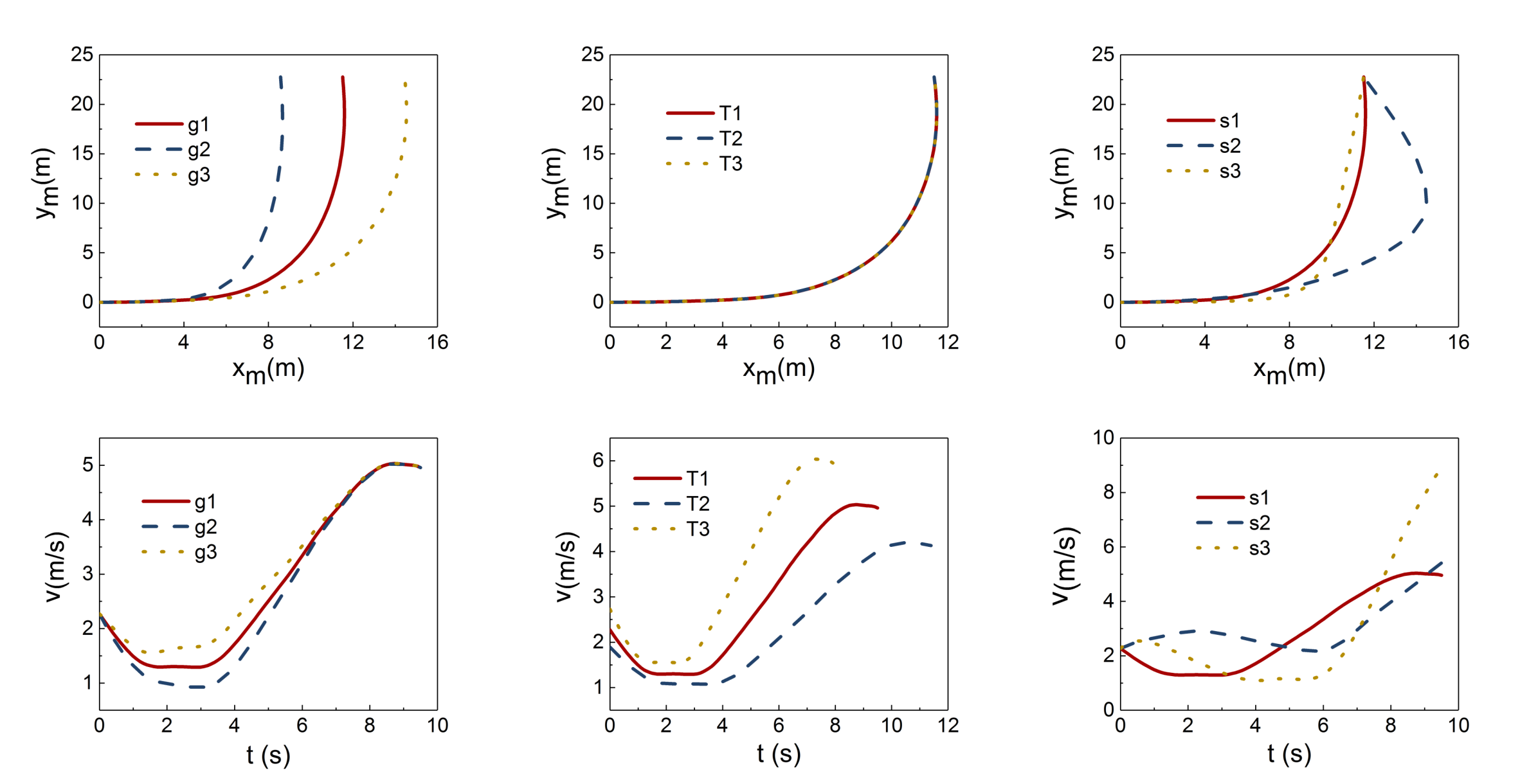}
\end{minipage}
\label{fig3(a)}
}

\subfloat[The type of lane changing MP]{
\begin{minipage}[t]{0.9\textwidth}
\centering
\includegraphics[scale=0.1865]{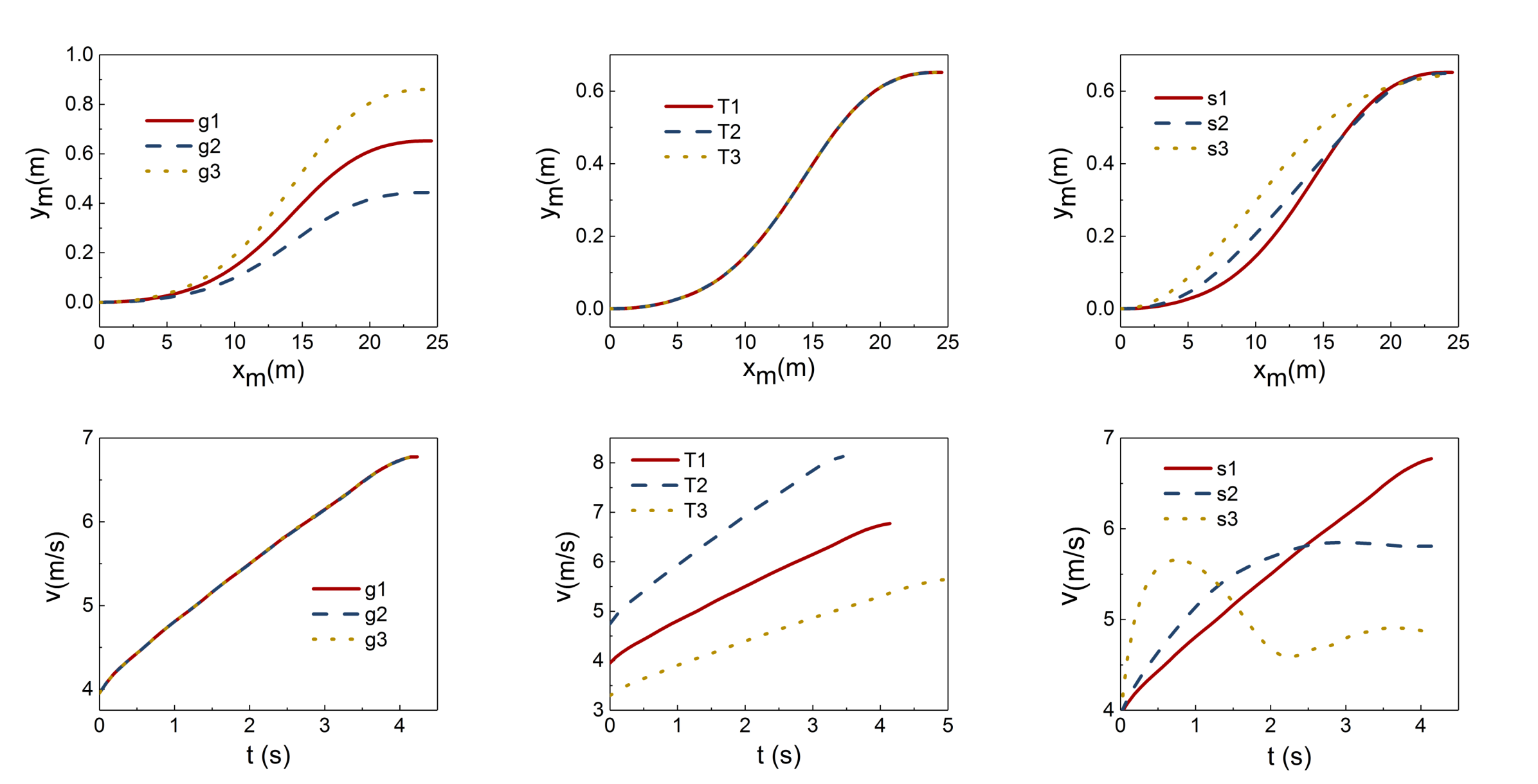}
\end{minipage}
\label{fig3(b)}
}
\caption{This figure shows the adjustment ability of the learned MPs duration regeneration. For the selected two types of MPs, we showed the corresponding adjustment results when the goal position $\boldsymbol{g}$, time duration $T$ and fine-tuning parameters $\boldsymbol{s}$ change respectively. During the adjustment process, when one of the parameters changed, the other two parameters remained unchanged. And the number of fine-tuning parameters $\boldsymbol{s}$ is set to $J=5$.}
\label{fig3}
\end{figure*}

\par The accuracy of the MP representation was evaluated by the average position deviation ${\Delta \bar d}$ and average velocity deviation ${\Delta \bar v}$, and the deviations were the absolute mean difference between all the demonstrated trajectories and the learned trajectories in the same type of MPs. We demonstrated the effect of different numbers of fine tuning parameter $\boldsymbol{s}$ (Eq.\ref{equ9}) on the representation accuracy in Fig.\ref{fig2}, and the corresponding average deviation values were presented in Table \ref{tab1}.

\par In addition, we also demonstrated the adaptability of the selected two MPs to the change of goal position $\boldsymbol{g}$, time duration $T$, and fine-tuning parameter $\boldsymbol{s}$ in Fig.\ref{fig3}.

\subsection{Evaluation of MPs Joining} 

In order to demonstrate the benefits of the proposed joining method, we applied the simple joining method as a baseline and evaluated the performance of the joining algorithm at high speed and low speed driving situations. And the evaluation indexes were the maximum acceleration $a_{max}$ and position deviation $\Delta d_k$. The maximum acceleration $a_{max}$ referred to the maximum value of the acceleration generated during the transition of the MP combination, and the position deviation $\Delta d_k$ referred to the minimum distance between the generated trajectory and the target points set by the initial condition.

\par
\begin{figure}
\centering
\subfloat[Low speed driving situation]{
\begin{minipage}[t]{0.45\textwidth}
\centering
\includegraphics[scale=0.1]{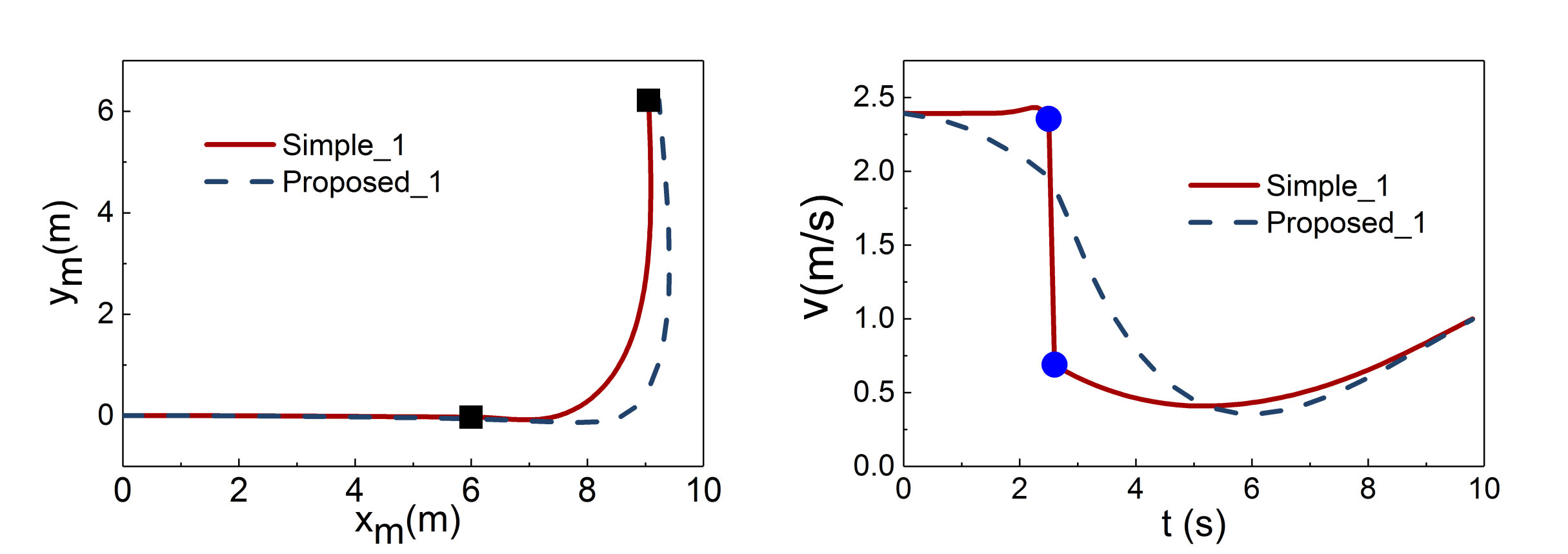} 
\end{minipage}
\label{fig4(a)}
}

\subfloat[High speed driving situation]{
\begin{minipage}[t]{0.45\textwidth}
\centering
\includegraphics[scale=0.1]{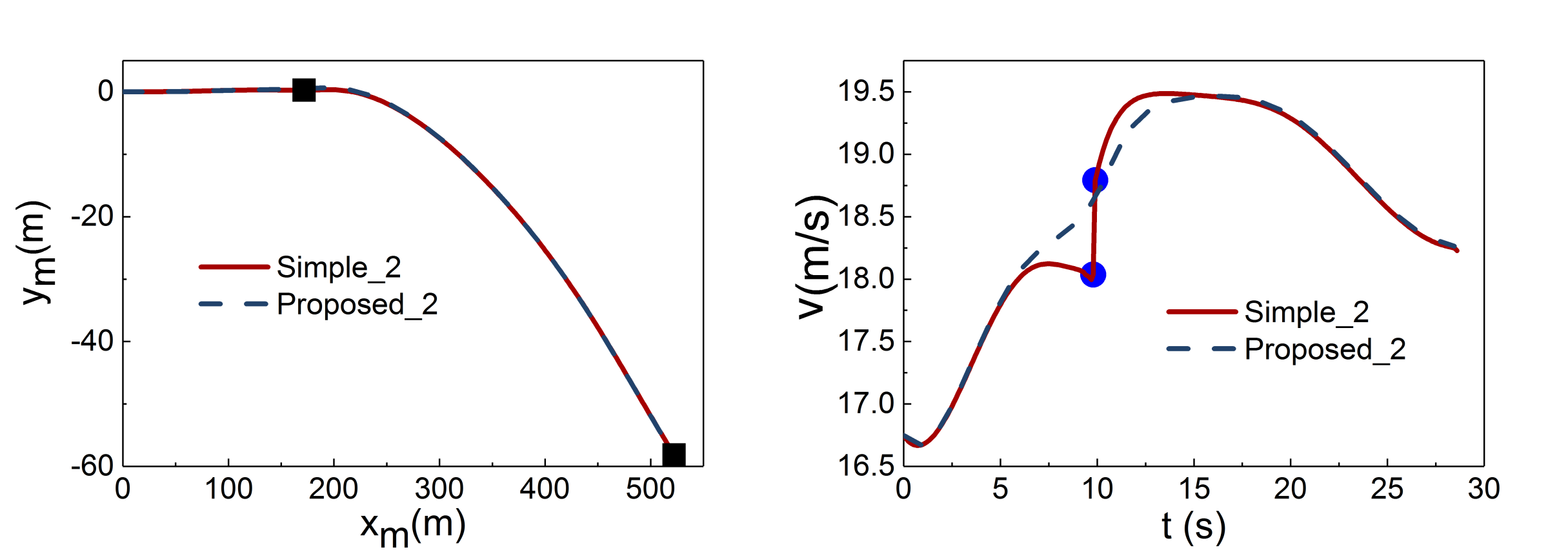} 
\end{minipage}
\label{fig4(b)}
}
\caption{This figure demonstrates the joining performance of MPs in low speed and high speed conditions. The black points were the target position points set by the initial condition. The blue points were the indications of the velocity at the end of the previous MP and the initial velocity of the next MP when connected via the simple joining method.}
\label{fig4}
\end{figure}

 The generation results of the MP sequence based on the two methods were shown in Fig.\ref{fig4}, and the evaluation indexes for the low-speed and high-speed situations were shown in Table \ref{tab2}.

\subsection{Discussion}

Through the SVD, we not only extracted the main trajectory shape features from multiple demonstration trajectories but also significantly reduced the number of trajectory shape adjustment parameters when the MPs were regenerated. In the original DMP representation algorithm, the trajectory shape was completely represented and adjusted by the parameter set $\boldsymbol{\omega}$ (Eq.\ref{equ4}). In order to obtain acceptable representation accuracy, the value of $N$ (Eq.\ref{equ4}) was usually set to 20. In our proposed method, we still used the $\boldsymbol{\omega}$ to complete the basic shape representation task but used $\boldsymbol{s}$ to complete the shape adjustment task (Eq.\ref{equ9}). Although the value of $N$ is still 20, the value of $J$ is greatly reduced. With this feature, we can fine-tune the shape of the trajectory with limited parameters without affecting the accuracy of the representation. 

\par From Fig.\ref{fig2} and Table \ref{tab1}, we can observe that for all the demonstration trajectories in the two selected types of MPs, the position deviation and the velocity deviation decreased with the increase of the value of $J$. However, the improvement of the representation accuracy was inconsistent for multiple demonstration trajectories in the same type. That is to say, the increase of the $J$ introduced more trajectory shape features in the chosen type, which improved the representation ability of the proposed method for different demonstration trajectories in the same type. In order to limit the number of shape adjustment parameters, we only selected some shape features, which also sacrificed the representation accuracy. However, this sacrifice in accuracy reduced the number of MPs shape adjustment parameters, and the reduction of adjustment parameters is of great significance for the future utilization of the MP library.

\par As for the MP regeneration as shown in Fig.\ref{fig3}, we retain the ability of the original DMP representation method to adapt to the changes in goal position and time duration, and add fine-tuning capabilities of the trajectory shape to further improve the adaptability of MPs to different motion planning tasks. The joining method proposed in this paper transforms the problem of generating MP sequences into a re-presentation problem of a single MP. By associating the attribute information of each independent MP in the sequence, the kernel function, the goal function, and canonical system are redefined and the re-representation task is completed. Since the MP sequence can be considered as a whole after re-representation, the transition problem between MPs is converted into a single MP regeneration problem. And we can easily achieve a smooth transition by utilizing the properties of modified DMP representation algorithm. From the results in Fig.\ref{fig4} and Table \ref{tab2}, we can find that although there were position deviations when tracking the set target points, the maximum acceleration of the proposed method in the joining process was only $5.8\%$ of the simple joining method in situation 1 and $4.6\%$ in situation 2. 

\section{Conclusion}

\begin{table}
 \centering
 \caption{\label{tab2} This table lists the maximum acceleration $a_{max}$ and position deviations $\Delta d_1$, $\Delta d_2$ during the joining process. $\Delta d_1$ corresponds to the first target point, $\Delta d_2$ corresponds to the second target point. The types listed in the table are consistent with the labels in Fig.\ref{fig4}.}
 \begin{tabular}{c c c c}
  \toprule
  Type & $\Delta d_1$(m) & $\Delta d_2$(m) & $a_{max}$($\rm{m/s\ ^2}$)\\ 
  \midrule
  Simple-1 & 0 & 0 & 16.65  \\
  Proposed-1 & 0.02 & 0.21 & 0.98  \\
  Simple-2 & 0 & 0 & 7.52  \\
  Proposed-2 & 0.32 & 0.07 & 0.34 \\
  \bottomrule
 \end{tabular}
\end{table}

This paper proposed a modified MP representation algorithm to deal with the need for MPs regeneration and joining in the automated vehicle motion planning system. The proposed method can represent the multiple similar demonstrated trajectories with acceptable precision and reduce the number of shape adjustment parameters during regeneration. This improvement is due to the introduction of SVD, which enables us to separate the basic shape parameters and fine-tuning adjustment parameters and to extract the main features from multiple demonstration trajectories of the same type. Besides, the proposed MP representation method also has the ability to extend from a single MP representation to a MP sequence representation and uses its smoothing ability in the regeneration process to avoid the velocity jump under the premise of ensuring accuracy. The proposed method is to further improve the applicability of the previously established motion primitive library by enhancing the adjustment ability and achieving a smooth transition. With the above two aspects of improvement, the established MP library can better meet the needs of the motion planning system and help to complete personalized driving tasks.

\par In future work, we will further enhance the representation capabilities of the MPs so that different driving styles can be incorporated into the representation algorithm in the form of adjustment parameters, and combine the established MP library with the existing motion planning algorithm.

\bibliographystyle{unsrt}
\bibliography{root}

\end{document}